\documentclass{article}

\usepackage{microtype}
\usepackage{graphicx}
\usepackage{subfigure}
\usepackage{booktabs} 

\usepackage{hyperref}

\usepackage[accepted]{icml2024}

\usepackage{amsmath}
\usepackage{amssymb}
\usepackage{mathtools}
\usepackage{amsthm}
\usepackage{natbib}
\usepackage[capitalize,noabbrev]{cleveref}

\theoremstyle{plain}

\theoremstyle{definition}

\theoremstyle{remark}

\usepackage[textsize=tiny]{todonotes}
\usepackage{twemojis}
\usepackage{color}

\begin{document}

\twocolumn[
\icmltitle{EvGGS: A Collaborative Learning Framework for Event-based Generalizable Gaussian Splatting}



\icmlsetsymbol{equal}{*}

\begin{icmlauthorlist}
\icmlauthor{Jiaxu Wang}{yyy}
\icmlauthor{Junhao He}{yyy}
\icmlauthor{Ziyi Zhang}{yyy}
\icmlauthor{Mingyuan Sun}{yyy2}
\icmlauthor{Jingkai Sun}{yyy}
\icmlauthor{Renjing Xu}{yyy}
\end{icmlauthorlist}

\icmlaffiliation{yyy}{Function Hub, Hong Kong University of Science and Technology, Guangzhou, China}
\icmlaffiliation{yyy2}{Faculty of Robot Science and Engineering, Northeastern University, Shenyang, China}

\icmlcorrespondingauthor{Renjing Xu}{renjingxu@ust.hk}

\icmlkeywords{Machine Learning, ICML}

\vskip 0.3in
]



\printAffiliationsAndNotice{}  

\begin{abstract}
Event cameras offer promising advantages such as high dynamic range and low latency, making them well-suited for challenging lighting conditions and fast-moving scenarios. However, reconstructing 3D scenes from raw event streams is difficult because event data is sparse and does not carry absolute color information. To release its potential in 3D reconstruction, we propose the first event-based generalizable 3D reconstruction framework, called EvGGS, which reconstructs scenes as 3D Gaussians from only event input in a feedforward manner and can generalize to unseen cases without any retraining. This framework includes a depth estimation module, an intensity reconstruction module, and a Gaussian regression module. These submodules connect in a cascading manner, and we collaboratively train them with a designed joint loss to make them mutually promote. To facilitate related studies, we build a novel event-based 3D dataset with various material objects and calibrated labels of grayscale images, depth maps, camera poses, and silhouettes. Experiments show models that have jointly trained significantly outperform those trained individually. Our approach performs better than all baselines in reconstruction quality, and depth/intensity predictions with satisfactory rendering speed. Code and Dataset are demonstrated \textcolor{blue}{https://github.com/Mercerai/EvGGS/}

\end{abstract}
    
\vspace{-0.5cm}
\section {Introduction}
\label{sec. intro}
3D reconstruction has played a critical role in computer vision communities and is vital in many applications, e.g. robotics, VR/AR, and graphics. Recently, several works have proposed promising approaches that can reconstruct high-fidelity 3D scenes from a moving RGB camera (to collect multiviews), such as Neural Radiance Field (NeRF) \cite{mildenhall2021nerf} and 3D Gaussian Splatting \cite{kerbl20233d}. However, conventional RGB cameras suffer from severe motion blurs when the moving speeds of cameras are fast and cannot be used in extreme lighting/dark environments due to their low dynamic ranges. The bio-inspired event cameras independently respond to log-intensity changes for each pixel asynchronously, instead of measuring absolute intensity synchronously at a constant rate, like in standard cameras. These unique principles contribute to multiple advantages of event cameras: high dynamic ranges, low latency, and high temporal resolution. 

Most existing 3D vision methods merely focus on standard cameras and do not provide event-based solutions because the output of event streams is very different from ordinary images, which are composed of the polarity, pixel location, and time stamp, occurring only at a sparse set of locations. A few studies \cite{rudnev2023eventnerf, hwang2023ev} attempt to combine event cameras with NeRF, but their rendering results struggle with blurred edges and boundaries, and soft fogs often exist in front of the camera lens. The reason is NeRF encodes scenes in continuous networks, thereby cannot effectively fit discontinuities and empties which are common in event representations. More recently, 3DGS introduced a novel representation that formulates the scene as 3D Gaussians with learnable parameters including color, opacity, and covariance. 3DGS enables more photo-realistic renderings with less memory cost and faster rendering speeds. Likewise, 3DGS only reconstructs a scene from per-scene optimization. 

Furthermore, the above-mentioned event-based neural reconstruction methods require per-scene optimization, and cannot generalize to unseen scenes. In contrast, some works \cite{lin2022efficient, zheng2023gps} have investigated the generalizable NeRF and 3DGS for RGB frames. However, no work adapts these approaches to event data at present. This is because most generic NeRFs rely on image-based rendering, they perform spatial interpolation across nearby views to the target view, whereas the event stream does not contain rich information to interpolate novel views. This work attempts to reconstruct 3DGS from raw event data in a feedforward manner, enabling it to generalize to unobserved scenarios without re-optimization. 

On the other hand, depth estimation and intensity recovery from raw event streams are still challenges \cite{jianguo2023stereo, hidalgo2020learning}. In standard cameras, stereo depth estimation relies on finding corresponding points in different camera views to triangulate depth. However, the absence of clear correspondence between events in different views makes stereo-matching difficult. Monocular depth prediction often relies on color information which event cameras do not include. While the recovery of intensity images is effective when interpolating between the given intensity images \cite{wangeslnet}, the performance dramatically degrades when only the event stream is available \cite{rebecq2019high}. In this work, we collaboratively train these subtasks under the 3DGS framework. The 3D-aware learning paradigm could improve the performances of subtasks because they mutually benefit from each other and in turn feedback on the quality of 3DGS reconstruction.

The contributions of this paper are summarized as follows: \textbf{(1)} We first propose the pure event-based, generalizable 3DGS framework (EvGGS), which faithfully reconstructs 3D scenes as 3D Gaussians from raw event streams and generalizes to various unseen scenarios. The proposed method outperforms existing event-based methods.

\textbf{(2)} We propose an end-to-end collaborative learning framework to jointly train event-based monocular depth estimation, intensity recovery, and 3D Gaussian reconstruction by connecting these modules in a cascading manner. Experiments show that the 3D-aware training framework yields better results than those individually trained models. 

\textbf{(3)} To facilitate related studies, we establish a novel event-based 3D dataset (Ev3DS) with varying material objects and well-calibrated frame, depth, and silhouette groundtruths.

\section{Related Work}
\label{sec. related work}
\subsection{Neural 3D reconstruction}
Traditional explicit representation methods include point cloud\cite{achlioptas2018learning}, mesh\cite{liu2020general}, and voxel\cite{lombardi2019neural,sitzmann2019deepvoxels}. However, they are limited by their fixed topological structure. As a solution, implicit representation has been proposed \cite{liu2020dist}, but these methods still require the input of surface features of the scene as prior.

NeRF\cite{mildenhall2020nerf} employs an MLP to reconstruct a scene and synthesizes images by volume rendering, but it requires a very long time for optimization. Recently, some studies \cite{cao2023hexplane, chen2022tensorf, barron2021mip, huang2023ponder, zheng2023pointavatar, lionar2021dynamic, chen2023neurbf} have combined implicit NeRF with explicit 3D representation to overcome its issues. \cite{fridovich2022plenoxels} and \cite{sun2022direct} store neural features into voxel grids rather than MLP to skip empty space. NeuMesh\cite{yang2022neumesh} distills the neural field into a mesh scaffold, enabling field manipulation with the mesh deformation. Ref-NeuS\cite{ge2023ref} model sign distance field by incorporating explicit reflection scores into NeRF. \cite{xu2022point} and \cite{wang2023learning} combine point clouds with NeRF to deliver better reconstruction quality. In contrast to NeRF, \cite{kerbl20233d} proposed the 3D Gaussian Splatting, which demonstrates remarkable performance in terms of rendering quality and convergence speed. 

Recently a few studies have attempted to directly apply the neural reconstruction methods to raw event streams. \cite{rudnev2023eventnerf, hwang2023ev, klenk2023nerf, wang2024physical} build similar pipelines which integrate the event generation model into NeRF. Nevertheless, these approaches still suffer from the various limitations we listed in Sec.\ref{sec. intro}. In this work, we first combine 3DGS with event-based reconstruction, improving the quality of reconstruction from pure event data.

\subsection{Generalizable neural reconstruction}
Either NeRF or 3DGS require per-scene optimization because they need gradient backpropagation to adjust their intrinsically scene-specific parameters. To address this, some works attempt to propose generalizable methods to construct a NeRF on the fly. MVSNeRF\cite{chen2021mvsnerf} and IBRNet\cite{wang2021ibrnet} achieve cross-scene generalization from only three nearby input views by building feature augmented cost volume. ENeRF\cite{lin2022efficient} utilizes a learned depth-guided sampling strategy to improve the rendering efficiency. NeuRay\cite{liu2022neural} implicitly models visibility to deal with occlusion issues. GPF \cite{wang2024gpf} proposes to fully utilize the geometry priors to explicitly improve the sampling and occlusion perception. Very recently, \cite{zheng2023gps} proposed the first generalizable 3D Gaussian framework for real-time human novel view rendering. However, all the above generic NeRF approaches only focus on RGB cameras, and the method for raw event data is still blank.

\subsection{Learning-based Event Depth and Image Estimation}
Estimating depth from events is challenging because event data only contains relative illumination changes, which are not suited to feature matching across views. \cite{hidalgo2020learning} yields a recurrent architecture to solve this task and show over $50\%$ improvement compared to traditional hand-crafted methods. EReFormer\cite{liu2022event} introduces a spatial fusion module and a gate recurrent transformer for temporal modeling to predict monocular depth. ASnet\cite{jianguo2023stereo} utilizes a group of adaptive weighted stacks to extract depth-related features.  \cite{brebion2023learning} fuses information from an event camera and a LiDAR. 	
Intensity image reconstruction from only event input has been another popular topic in event camera research\cite{cadena2021spade, paredes2021back, liu2023sensing}. E2VID\cite{rebecq2019high} introduced a ConvLSTM-based model, facilitating the recovery of high-dynamic video. FireNet\cite{scheerlinck2020fast} employs the GRUs to provide a more rapid and lightweight method for event-based video reconstruction. ET-Net\cite{weng2021event} employed a vision transformer to reconstruct videos from events. EVSNN\cite{zhu2022eventbased} proposes a hybrid potential-assisted spiking neural network to recover images from events efficiently. 

At present, both the two tasks from events still require further improvement. In this work, we collaboratively optimize the two tasks under the 3D Gaussian rendering framework to mutually promote their performance.

\section{Preliminary}
\label{sec. preliminary}
Since the proposed framework is related to event-based vision and 3D Gaussian, we give brief and basic knowledge about the two sides in this section. 
\vspace{-1.5mm}
\subsection{3D Gaussian Splatting}
\vspace{-1mm}
3DGS parameterize a 3D scene as a series of 3D Gaussian primitives, each has a mean ($\mu_k$), a covariance ($\sum_k$), an opacity ($\alpha_k$) and spherical harmonics coefficients ($\textbf{SH}_k$). These primitives parameterize the 3D radiance field of the underlying scene and can be rendered to produce novel views via Gaussian rasterization. To facilitate optimization by backpropagation, the covariance matrix can be decomposed into a rotation matrix ($\textbf{R}$) and a scaling matrix ($\textbf{S}$):
\vspace{-1.5mm}
\begin{equation}
    \Sigma = RSS^TR^T
    \vspace{-1mm}
\end{equation}
Assuming the camera trajectory is known, the projection of the 3D Gaussian to 2D image plane can be described by the view transformation ($\textbf{W}$) and the projection transformation. To maintain the linearity of the projection, the Jacobian of the affine approximation $\textbf{J}$ of the projective transformation is applied, as in:
\vspace{-1.5mm}
\begin{equation}
    \Sigma^{'} = JW\Sigma W^TJ^T
    \vspace{-1mm}
\end{equation}
where the $\Sigma^{'}$ is the projected 2D covariance. The $\alpha$-blend is used to compute the final color of each pixel.
\vspace{-1.5mm}
\begin{equation}
    C = \sum_{i\in \mathcal{N}}c_i\alpha_i\prod_{j=1}^{i-1}(1-\alpha_j)
    \vspace{-1mm}
\end{equation}
The above parameters can be summarized in the following. $\mu$ is the position of a primitive $\mu \in R^3$. The rotation matrix is parameterized by a quaternion $q \in R^4$. The scale factor refers to the anisotropy stretching $s \in R^3$. The 2D opacity $\alpha \in [0,1]$ is computed by $\alpha_i(x) = o_iexp(-\frac{1}{2}(x-\mu_i)^T{\Sigma}_{i}^{T}(x-\mu_i))$ where the $\mu$ and $variance$ are the 2D-projected mean and variance of 3D Gaussians. The color is defined by $\textbf{SH}$.
\vspace{-3mm}
\subsection{Event Representation}
Events ($e_k=(\textbf{u}_k,t_k,p_k)$) occur asynchronously at pixel $\mathbf{u}_k=(u, v)$ with micro-second timestamp $t_k$. The brightness changes determine the polarity ($p \in \{+1, -1\}$). An event at time $t_k$ can be triggered following the equation:
\vspace{-1.5mm}
\begin{equation}
   \Delta L_k(\textbf{u}) = \sum_{e_i\in\Delta t_k}p_iC
   \vspace{-2mm}
\end{equation}
\vspace{-0.85mm}
where $L$ denotes the logarithmic frame ($L(t)=log(I(t)$) and $C$ refers to the constant threshold. Thus, if the $C$ is given, we could accumulate the events for a given period $\Delta t$ to obtain the log brightness difference in a specific pixel. To process the event stream synchronously, we encode the events in $\Delta t$ in a spatial-temporal voxel grid. The duration $\Delta t$ is discretized into $B$ temporal bins. Each event trilinearly contributes to its near voxels by its polarity, as stated in:
\vspace{-2mm}
\begin{equation}
    E(u,v, t_n) = \sum_{i}p_i\max(0,1-|t_n-t_i^*|)
    \vspace{-2mm}
\end{equation}
where $t_i^*$ is determined by the number of bins and is normalized to 0 to 1 $t_i^* = \frac{B-1}{\Delta t}(t_i - t_0)$. Following \cite{scheerlinck2020fast}, we set $B=5$ in our experiments. 

\section{Methodology}
\label{sec. Method}
\begin{figure*}[h]
    \centering
    \includegraphics[width=1\linewidth]{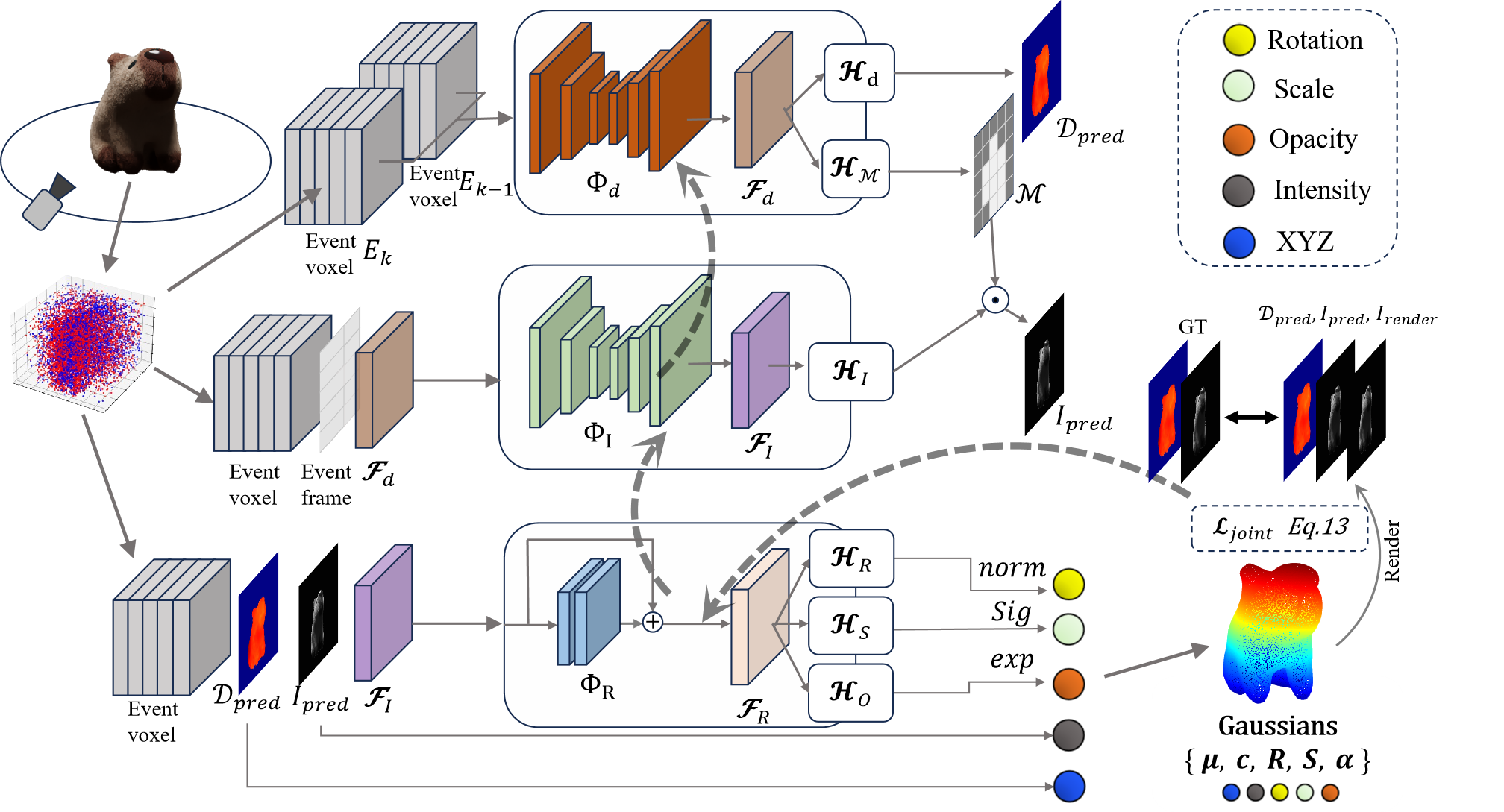}
    \vspace{-9mm}
    \caption{\textbf{Overview of EvGGS.} Given a 360-degree event stream and target viewpoints. we select two segments of event spatial-temporal voxels from consecutive moments as inputs. For each source view, we employ two submodules to extract the depth and intensity information, which serve as the 3D position and color maps. Another module aims to infer other 3D Gaussian parameters. The feature and output of the three modules are hierarchically bridged, facilitating a smooth backpropagation through joint training.}
    \label{fig: main pipeline}
    \vspace{-4mm}
\end{figure*}
Figure~\ref{fig: main pipeline} illustrates the whole pipeline of our proposed approach. The primary goal of our method is to reconstruct the 3D Gaussians of scenes in a feedforward manner from the given event stream captured by a moving event camera. The by-product of our method contains satisfactory depth and intensity prediction models. The proposed framework includes three main components: the depth and mask prediction module, the intensity reconstruction module, and the 3D Gaussian parameter regression module. We jointly train them to enable them to benefit from other tasks. The 360-degree event stream is divided into 201 segments corresponding to the 201 grayscale images for each scene. A dense depth map and a corresponding intensity map are predicted for each event segment. It is noted that the normal event camera only detects brightness changes rather than recognizing colors. Therefore, the event-based 3DGS only produces the intensity parameter $ I \in R^1$ instead of the spherical harmonics coefficients $\textbf{SH}$. Next, the Gaussian regressor predicts other parameters. The depth map and associated parameters are unprojected to the 3D space. As shown in Figure~\ref{fig: main pipeline}, three main modules are hierarchically linked in both feature and output spaces. The gradient can be efficiently back-propagated through the pipeline, thus allowing for efficient joint optimization. 

\subsection{Event-based Monocular Depth Estimation Module}
The depth estimation module takes two segments of spatial-temporal event voxel grid $E_k$ and $E_{k-1}$ from consecutive moments as inputs. We let the module predict the normalized log disparity map. The final depth value can be converted from the predicted disparity:
\vspace{-2mm}
\begin{equation}
    D_{pred}=exp(D_{max}\odot Sig(Disp)))
    \vspace{-1mm}
\end{equation}
where $D_*$ refers to the associated disparity map, $Sig$ denotes the sigmoid activation to ensure the output value belongs to $(0,1)$. We additionally attempt to directly regress normalized depth value and similar results are observed in the final experiments. For simplicity, we do not specifically distinguish the two terms in the rest of the paper.

We implement this module with a dense UNet. Detailed network architectures are shown in the \textbf{Appendix}. The output of the UNet is fed to two output heads to predict the normalized depth and the foreground mask. Moreover, the output feature volume of the UNet is also maintained to pass to the next intensity reconstruction module. The foreground mask is multiplied with all 3D Gaussian parameter maps to filter out the useless and empty backgrounds. The whole process can be depicted in the following:
\vspace{-1mm}
\begin{equation}
\begin{aligned}
&\mathcal{F}_d=\Phi_d(E(u,v,t),E(u,v,t-1))\\
&\mathcal{D}, \mathcal{M} = Sig(\mathcal{H}_d(\mathcal{F}_d)), Sig(\mathcal{H}_m(\mathcal{F}_d))\\
&Disp = \mathcal{D} \odot \mathcal{M}
\label{eq. depth and mask prediction}
\end{aligned}
\vspace{-1.5mm}
\end{equation}
where the $\Phi_d$ refers to the depth UNet and $\mathcal{F}_d \in R^{H \times W \times 32}$ refers to the 32-dimensional output feature volume which can be decoded to normalized depth maps and mask maps by the corresponding heads $\mathcal{H}_d$ and $\mathcal{H}_m$. Next, the $\odot$ refers to the element-wise multiplication. 

\subsection{Intensity Reconstruction Module}
The intensity reconstruction module aims to offer the color properties of 3D Gaussians (as for the event version, the color denotes intensity.) This module receives the event voxel grid, accumulated event frame, and the depth feature volume from the previous module as input ($\mathcal{F}_d$ in Equation~\ref{eq. depth and mask prediction}) to utilize the geometry awareness to assist appearance recovery. The network architecture follows the depth estimation module, and is UNet-like as well.  
\vspace{-1mm}
\begin{equation}
    \mathcal{F}_I = \Phi_I(\mathcal{F}_d \oplus E(u,v,t) \oplus F(u,v))
    \label{eq. intensity feature}
    \vspace{-1mm}
\end{equation}
where $\Phi_I$ represents the UNet network with a similar architecture as that in module 1. The $\oplus$ denotes the concatenation operation. Moreover, $F(u, v) \in R^{H\times W\times 3}$ represents the accumulated event frame, which is produced by accumulating events at the same pixel location together, and we repeat the operation three times for different polarity combinations including positive, negative,  positive and negative, respectively, and concatenate them along the channel dimension because the event frame contains rich boundary information which helps recover dense intensity maps. The final reconstructed intensity map can be obtained by 
\vspace{-1mm}
\begin{equation}
    \mathcal{I}_{pred} = \mathcal{M} \odot Sig(\mathcal{H}_I(\mathcal{F}_I))
    \label{eq. intensity pred}
    \vspace{-1mm}
\end{equation}
in which $\mathcal{M}$ is the predicted foreground mask in Equation~\ref{eq. depth and mask prediction}. The cascaded connection guarantees that geometric priors are taken into account when the module deduces appearance. 

\begin{figure*}
    \centering
    \includegraphics[width=1\linewidth]{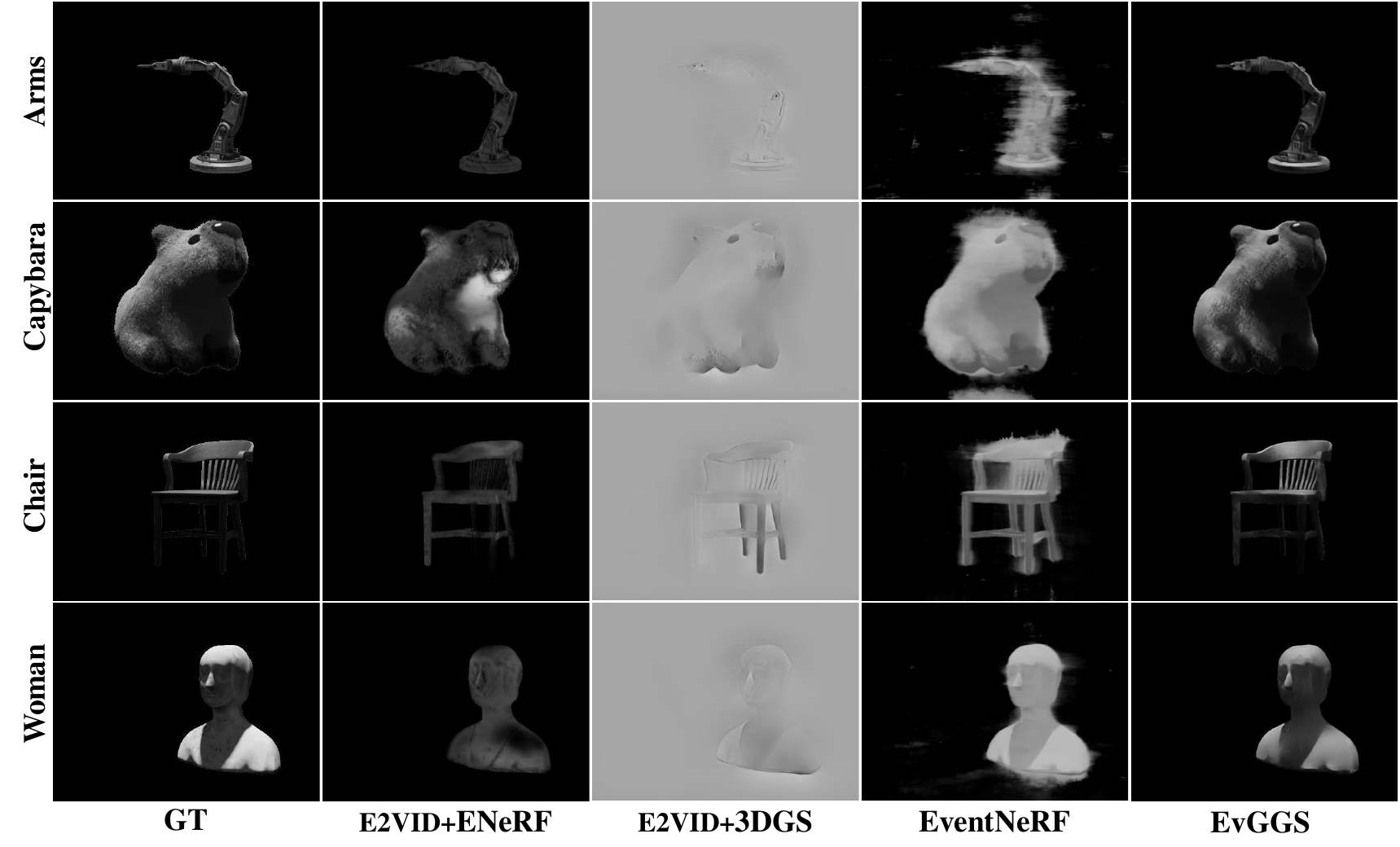}
    \vspace{-8mm}
    \caption{Qualitative comparison of ours and other event-based 3D methods in novel view synthesis. }
    \vspace{-4mm}
    \label{fig: 3d}
\end{figure*}
\vspace{-1mm}
\subsection{Gaussian Parameter Regression}
\vspace{-1mm}
As stated in Section~\ref{sec. preliminary}, the 3D Gaussian includes 5 independent parameters $\mu, \textbf{R}, \textbf{S}, \alpha, c$. The first two modules generate the 3D location and intensity, and then the regressor indicated in this subsection aims to formulate the rest parameters, i.e. scale, rotation, and opacity. This module is a residual block with two convolutional layers. 
\begin{equation}
    \mathcal{F}_R = \Phi_R(\mathcal{D}_{pred} \oplus \mathcal{F}_I \oplus E(u,v,t))
\label{eq. regressor}
\end{equation}
where the $\mathcal{F}_I$ represents the output feature volume of the intensity module in Equation~\ref{eq. intensity feature} and $\mathcal{D}_{pred}$ is the predicted depth. The $\mathcal{R}_R$ is decoded into different Gaussian parameters with corresponding activation functions to constrain the value range.
\begin{equation}
\begin{aligned}
&\textbf{R} = norm(\mathcal{H}_r(f_R)) \\
&\textbf{S} = exp(\mathcal{H}_s(f_R)) \\
&\alpha = Sig(\mathcal{H}_o(f_R)) 
\label{sec. regression gs param}
\end{aligned}
\end{equation}
in which the $\mathcal{H}_r$, $\mathcal{H}s$, and $\mathcal{H}_o$ represent the corresponding decoder heads for different Gaussian parameters. $\textbf{R} \in R^{H \times W \times 4}, \textbf{S} \in R^{H \times W \times 3}, \alpha \in R^{H \times W \times 1}$. The predicted parameter maps have the same spatial resolution as the original input event voxel grid, the mask $\mathcal{M}$ in Equation ~\ref{eq. depth and mask prediction} is also used to filter invalid regions, $\textbf{R}_{pred}=\mathcal{M}*\textbf{R}$, $\textbf{S}_{pred}=\mathcal{M}*\textbf{S}$, $\alpha=\mathcal{M}*\alpha$. Moreover, the $\mathcal{D}_{pred}$ can be unprojected from pixel space to 3D space by giving the camera pose matrix $P \in R^{4 \times 4}$ and intrinsic matrix $K \in R^{3 \times 3}$ to obtain the parameter $\mu$, as stated in Equation ~\ref{eq. unprojection depth}.
\begin{equation}
    \mu = P \cdot K^{-1} \cdot (u,v,\mathcal{D}_{pred}(u,v))
    \label{eq. unprojection depth}
\end{equation}
Likewise, the predicted intensity $\mathcal{I}_{pred}$ serves as the Gaussian parameter $c$. 

\subsection{Training Strategy}
We jointly optimize the three modules by the differentiable rendering pipeline in an end-to-end manner. The framework finally renders intensity images that can be used to compute losses. We hierarchically bridge the three modules by linking their feature and output layers. Thanks to the hierarchical linkage, the gradient can smoothly backpropagate through the pipeline. Improved geometries provide better contextual information about the spatial relationships between different parts of an image and contribute to a better semantic understanding of the image. This understanding helps the model differentiate between different objects, regions, and surfaces, leading to more accurate texture reconstruction. In contrast, better texture reconstruction implies that fine details on surfaces are captured accurately. This detailed information is crucial for depth prediction, especially in regions with complex structures or intricate surfaces. Overall, better geometry and texture simultaneously improve the quality of the reconstructed 3D Gaussians. Due to the above analysis, multitasks in the collaborative learning framework mutually promote and benefit from each other. 

To mitigate the optimization complexity, we first pretrain the depth prediction module by $\mathcal{L}_1$ loss. In addition, We jointly train the whole pipeline according to the below Equation
\vspace{-1mm}
\begin{equation}
    \mathcal{L}_{joint}=argmin_{\phi, \theta, \eta}(\lambda_1 L_{I_\theta} + \lambda_2 L_{D_\phi} + \lambda_3 L_{R_{\phi, \theta,\eta}})
    \label{eq. total loss function}
    \vspace{-0mm}
\end{equation}
where $\phi, \theta, \eta$ corresponds to parameters of depth, intensity, and regressor modules. $\lambda_1, \lambda_2, \lambda_3$ are coefficients to balance the loss magnitudes. We set 0.2, 0.2, and 0.6 respectively throughout all experiments. In detail, the three losses are described as follows:
\begin{equation}
    \mathcal{L}_{I_\theta}=\beta_1 \mathcal{L}_2(I_\theta, I_{gt}^s) + \beta_2 L_p(I_\theta, I_{gt}^s)  
\end{equation}
\begin{equation}
    \mathcal{L}_{D_\phi}=\mathcal{L}_1(D_\phi, I_{gt}^s)
\end{equation}
\begin{equation}
\begin{aligned}
    \mathcal{L}_{R_\eta}&= \beta_1 \mathcal{L}_2(R_\eta(I_\theta, D_\phi), I_{gt}^t) \\
                         & + \beta_2 L_p(R_\eta(I_\theta, D_\phi), I_{gt}^t)
\end{aligned}
\end{equation}
In the above three loss equations, the superscripts $s$ and $t$ denote the source view and target view respectively. $\mathcal{L}_p$ is the perceptual loss \cite{zhang2018perceptual}. $\beta_1, \beta_2$ aim to balance the $\mathcal{L}_1$ and perceptual loss, we constantly set them to 0.8 and 0.2 for all situations. $I_\theta, D_\phi$ are the predictions of the first two modules at the source views. $R_\eta(I_\theta, D_\phi)$ represents the 3D Gaussian parameter regression and rasterization projection to the target view based on the source view predictions. In the inference stage, only the raw event stream is required to be the input. 
\vspace{-0.25cm}
\section{Experiments}
\label{sec. experiments}
\subsection{Event-based 3D Dataset}

\textbf{Dataset}
Existing event-based 3D datasets such as \cite{rudnev2023eventnerf, zhou2018semi} only contain a limited number of objects and lack high-quality intensity, depth, and mask groundtruths because they mainly concentrate on single scene reconstruction or sparse vision tasks. To fill the current gaps in the community, we establish a full event-based 3D dataset including completed labels, referred to as Ev3DS. The dataset includes a wide variety range of materials. There are 64 objects for training and 15 objects The dataset is constructed and rendered via Blender, it encompasses a multitude of photo-realistic objects, characterized by their complex and varied geometric structures and texture information. We have harnessed VisionBlender\cite{cartucho2020visionblender} to gather dense depth, mask, and pose information. In each scene, events are generated by a virtual event camera that orbits around the origin of the object in space. We employ V2E\cite{hu2021v2e} to generate synthetic event streams maintaining default noise configurations. Additionally, to verify the robustness of the proposed method, we establish and release a novel realistic dataset utilizing the event camera DVXplore for further faithful evaluations, referred to as Ev3D-R. The real-world dataset is essential because the real event camera will raise noises and be more sensitive to illumination changes than synthetic event data. We evaluate the scalability and generalization of the proposed methods on the realistic event data. A detailed introduction to Ev3D-R can be found in the \textbf{Appendix}.

\textbf{Metrics}. We evaluate the following three subtasks including depth estimation, intensity recovery, and novel view synthesis. Similar to previous works, we evaluate the absolute relative error, mean absolute error, square relative error, and root mean square error as metrics for the depth predictions in foreground regions. In addition, we evaluate PSNR, SSIM, and LPIPS for the intensity reconstruction and the novel view synthesis tasks as well. We here argue that Ours$_i$ and Ours$_j$ denote our models as independently trained and jointly trained respectively. Due to the page limit, more results and videos can be seen in the \textbf{Appendix} and Supplementary Material. 
\begin{figure*}[h]
    \centering
    \includegraphics[width=\linewidth]{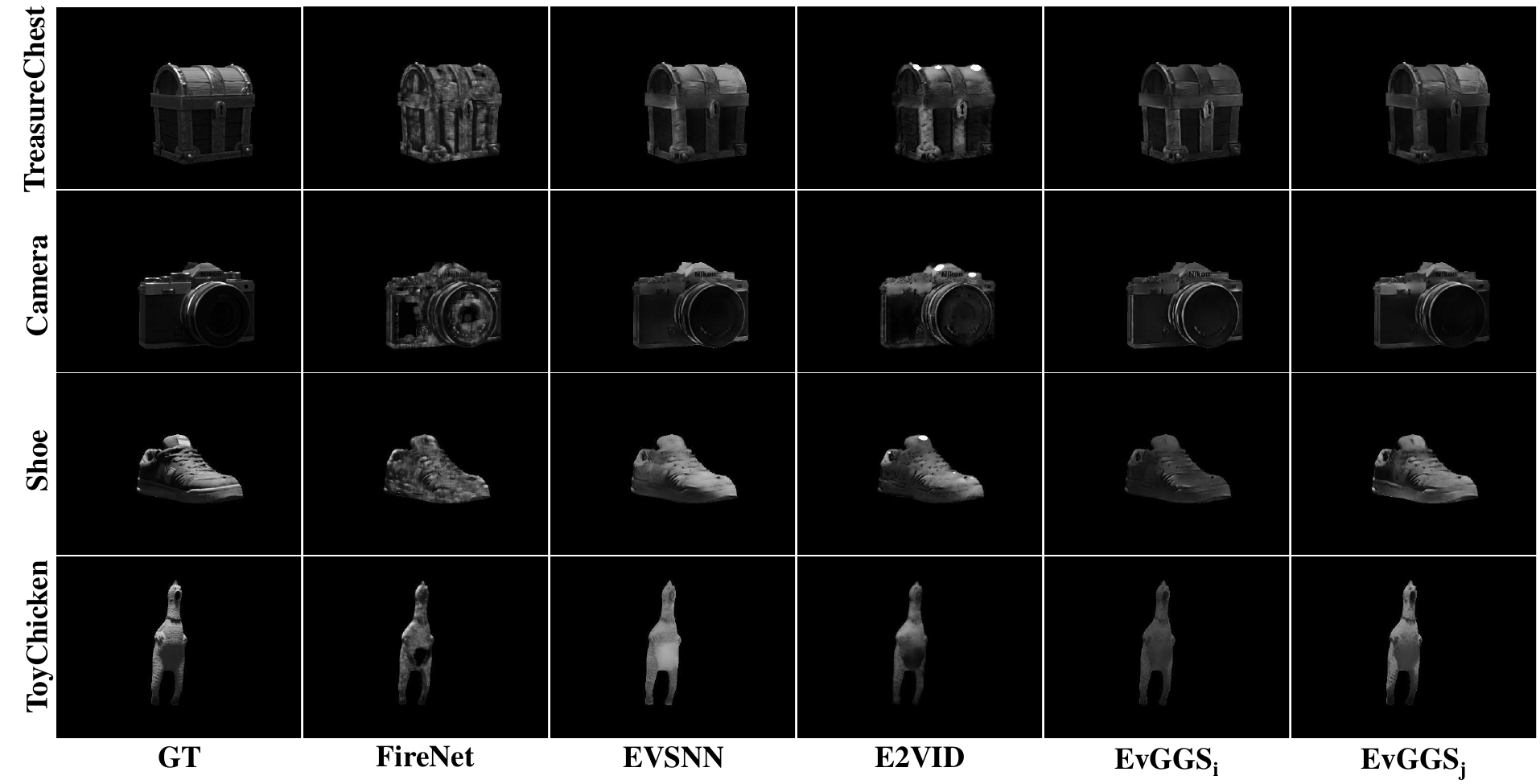}
    \vspace{-8mm}
    \caption{Qualitative comparison of ours and other intensity reconstruction methods.}
    \vspace{-6mm}
    \label{fig: Intensity}
\end{figure*}
\subsection{Performance of Neural Reconstruction}
\textbf{Baselines}. In this section, we evaluate the performance of the proposed method on the novel view synthesis. We set three experimental settings for fair comparisons including EventNeRF \cite{rudnev2023eventnerf}, E2VID+3DGS (E3DGS), and E2VID+ENeRF\cite{lin2022efficient} (EENeRF). EventNeRF is the up-to-date pure event-based NeRF method that requires per-scene training while our method can generalize to unseen scenes. E2VID+3DGS denotes that we first recover videos from events by the E2VID method then we use the reconstructed images to optimize the 3DGS representations. As our approach is a generic pipeline, we set E2VID+ENeRF as the generalizable reconstruction baseline in which ENeRF is a recently prevailing generalizable NeRF method, which requires source images to interpolate the target views and renders based on a built-in depth estimator. We retrain the ENeRF by the provided image groundtruth at first. Then we use E2VID to recover the intensity frame from events and input them to the ENeRF model for synthesizing novel images.   
\begin{figure*}[h]
    \centering
    \includegraphics[width=1\linewidth]{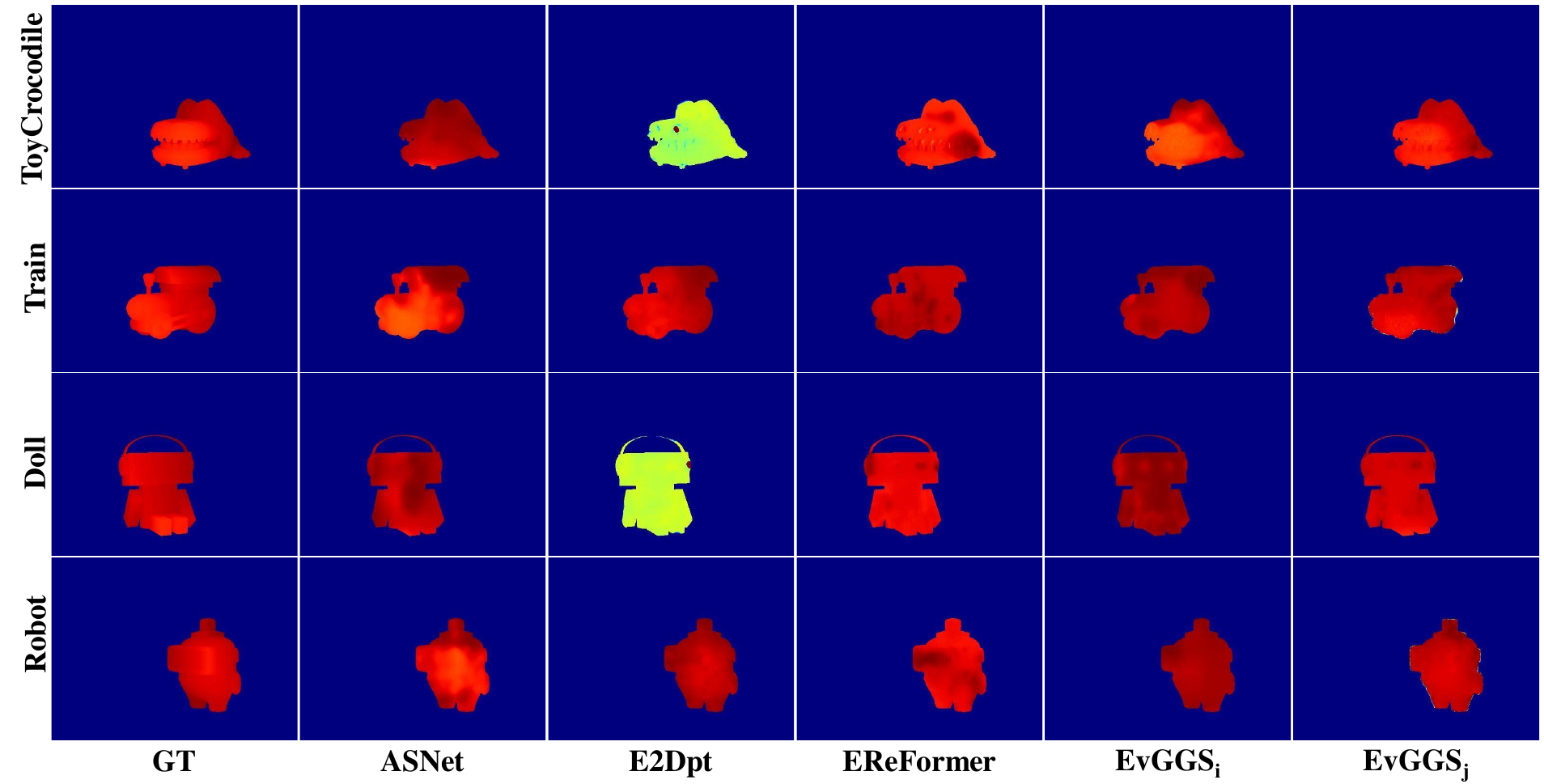}
    \vspace{-8mm}
    \caption{Qualitative comparison of ours and other depth estimation methods.}
    \vspace{-4mm}
    \label{fig: Depth}
\end{figure*}
Table~\ref{neural_table} reports all metrics to quantitatively compare the results, which indicates that our method achieves the best performance. Gen. means whether the corresponding methods can generalize to unobserved scenes. Real-time means the corresponding methods can render several images within 1 second. More analysis of this point is in the \textbf{Appendix}. Fig.~\ref{fig: 3d} presents the qualitative results. 

Compared to E2VID+ENeRF, the success can be attributed to the collaborative training that further improves the quality of depth and intensity prediction even though the performance of the intensity reconstruction module and E2VID is identical when trained individually. Compared to EventNeRF, EventNeRF suffers from soft fogs because NeRF encodes the scene into a continuous network, which greatly affects the quality of texture and geometric reconstruction of the objects. Our process effectively solves this issue. Compared to E2VID+3DGS, the reconstruction quality of E2VID+3DGS entirely depends on the quality of the intensity reconstruction module. 

\subsection{Quality of Intensity Reconstruction}
\textbf{Baselines}. We overall evaluate the quality of the intensity reconstruction in our framework. In this section, we select three popular image recovery algorithms, i.e. E2VID \cite{rebecq2019high}, FireNet \cite{scheerlinck2020fast}, and EVSNN \cite{barchid2023spiking}. They receive raw event data as input and reconstruct corresponding intensity maps. Moreover, we also independently train our intensity module as a baseline by using a fixed depth module to provide the $\mathcal{F}_d$ in Equation \ref{eq. intensity feature}. E2VID and FireNet rely on the recurrent convolution structure while EVSNN is built upon the spiking neural network. It is noted that the first three baselines recover videos via a recurrent mechanism, they need the last state to be an additional input, whereas our module directly infers the corresponding images from a segment of events. Therefore, they have to start to reconstruct from the first frame while ours can be reconstructed from arbitrary timestamps. 
We present the qualitative comparisons in Table \ref{intensity_table}. Figure~\ref{fig: Intensity} additionally shows some qualitative results.
\begin{table}
\caption{Qualitative Comparisons of neural reconstruction.}
\renewcommand{\arraystretch}{1}
\centering
\setlength\tabcolsep{1.52mm}{
\begin{tabular}{lcccc}
\hline\hline
Methods  & EventNeRF & E-ENeRF   & E3DGS   & EvGGS   \\ \hline
PSNR↑    & \underline{24.62}     & 23.86   & 19.19   &\textbf{27.95}  \\
SSIM↑    & \underline{0.945}     & 0.933   & 0.814   &\textbf{0.968}\\
LPIPS↓   & 0.072     & \underline{0.066}   & 0.119   &\textbf{0.045} \\
Gen.     & \twemoji{check mark} & \twemoji{check mark} & \twemoji{multiply} & \twemoji{check mark}\\
Real-time & \twemoji{multiply} & \twemoji{check mark} & \twemoji{check mark} & \twemoji{check mark}\\
\hline\hline
\end{tabular}}
\label{neural_table}
\vspace{-6mm}
\end{table}

\begin{table}
\caption{Qualitative Comparisons of intensity reconstruction.}
\renewcommand{\arraystretch}{1}
\centering
\setlength\tabcolsep{0.94mm}{
\begin{tabular}{lccccc}
\hline\hline
Methods    &FireNet &E2VID   &EVSNN  &EVGGS$_i$ & EVGGS$_j$  \\ \hline
PSNR↑    &25.56 &27.78 &\underline{27.91} &26.94 &\textbf{29.18}  \\
SSIM↑      &0.939 &0.963 &0.952 &\underline{0.957} &\textbf{0.969}\\
LPIPS↓      &0.056 &0.0567 &0.0397 &\underline{0.0367} &\textbf{0.0324} \\
\hline\hline
\end{tabular}}
\label{intensity_table}
\vspace{-6mm}
\end{table}
By comparing the results with the other three methods, it can be observed that our joint training strategy achieves superior performance in reconstructing complex textures. It effectively reconstructs the contrast that is close to the groundtruths and performs excellently in the local details. Compared to our independent training case, it can be seen that without the assistance of collaborative training, our intensity reconstruction module cannot reconstruct the contrast of the scene. Although it also reconstructs clearer textures, the reconstructed intensity images are still darker than the groundtruths.
\vspace{-1.5mm}
\subsection{Quality of Depth Estimation}
\textbf{Baselines}. In this section, we compare the depth estimation modules with four different baselines that are ASNet \cite{jianguo2023stereo}, EReFormer \cite{liu2022event}, E2Dpt \cite{hidalgo2020learning}, and our independent training strategy. ASNet is a stereo depth estimator that we give events of the target view and its nearest view for prediction. The rest are monocular depth estimators and we follow their original event representations as input. The results are depicted in Table \ref{depth_table}.

\begin{table}
\caption{Qualitative Comparisons of depth estimation. We magnified all metrics by a factor of 1000.}
\renewcommand{\arraystretch}{1}
\centering
\setlength\tabcolsep{0.95mm}{
\begin{tabular}{lccccc}
\hline\hline
Methods    &ASNet & E2Dpt &EReFormer &EvGGS$_i$ &EvGGS$_j$  \\ \hline
RMSE↓      & 2.87   & 2.86 & \underline{2.12} & 2.53 & \textbf{1.95} \\
Abs.rel↓      & 52.4   & 54.3 & \underline{46.2} & 51.5 & \textbf{39.4} \\
Sq.rel↓  & 4.38 & \underline{2.81} & 4.92 & 4.76 & \textbf{2.14}\\
\hline\hline
\end{tabular}}
\vspace{-6mm}
\label{depth_table}
\end{table}

We randomly select some examples to show in Figure \ref{fig: Depth}. Note that our joint training strategy achieved the best performance on test sets, especially in terms of the Abs. rel evaluation, where we achieved at least 14.7\% performance improvement compared to other baselines. As can be seen, our method can obtain finer-grained and more globally coherent dense depth maps across all test sets. Our method has a significant advantage when E2Dpt cannot predict the correct depth information. Compared to ASNet and EReFormer, our method achieves better results while using a more lightweight network structure, fully demonstrating the superiority of our joint training strategy.
\begin{table}
\caption{Ablation studies about different training strategies. PSNR, SSIM, and LPIPS are evaluated on the novel view synthesis. RMSE and Abs.rel are evaluated on the depth estimation.}
\renewcommand{\arraystretch}{1}
\centering
\setlength\tabcolsep{0.81mm}{
\begin{tabular}{lccccc}
\hline\hline
Methods     & PSNR↑  & SSIM↑ & LPIPS↓ & RMSE↓ & Abs.rel↓  \\ \hline
w/o Joint   & 27.04  & 0.953 & \underline{0.065}  & 2.53  & 51.5 \\
w/o Cascade & 26.51  & 0.934 & 0.068  & 2.51  & 51.6 \\
w/o $L_D$   & \underline{27.83}  & \underline{0.962} & 0.078  & 2.37  & 49.2 \\
w/o $L_I$   & 26.94  & 0.959 & 0.518  & \underline{1.98}  & \underline{41.6} \\
EvGGS        & \textbf{27.95}  & \textbf{0.968} & \textbf{0.045}  & \textbf{1.95}  & \textbf{39.4} \\
\hline\hline
\end{tabular}}
\vspace{-6mm}
\label{Ablation}
\end{table}

\subsection{Performance on Realistic Event Data}
We also evaluate our method and some baselines on Ev3D-R. Here all methods except the EvGGS-f are trained on the proposed synthetic dataset and directly tested on the Ev3D-R. It is observed that our method demonstrates the least sim2real gap and outperforms other baselines by a large margin, while others experience dramatic degeneration compared to the results on synthetic data. The EvGGS-f shows that the performance of the proposed approach can be further improved during fine-tuning.
\begin{table}
\caption{Qualitative Comparisons on Ev3D-R.}
\renewcommand{\arraystretch}{1}
\centering
\setlength\tabcolsep{.5mm}{
\begin{tabular}{lccccc}
\hline\hline
Methods  & E-ENeRF & FireNet  &EVSNN & EvGGS-g   & EvGGS-f   \\ \hline
PSNR↑    & 23.87   & 23.64   & 24.95 &26.77  &\textbf{27.84}  \\
SSIM↑    &0.866     & 0.833   & 0.643 &0.896  &\textbf{0.927}\\
LPIPS↓   & 0.271     & 0.267   & 0.020  &0.128 &\textbf{0.086} \\
\hline\hline
\end{tabular}}
\label{table: ev3dr}
\vspace{-8mm}
\end{table}
\renewcommand{\dblfloatpagefraction}{.9}
\begin{figure*}
    \centering
    \includegraphics[width=0.9\linewidth]{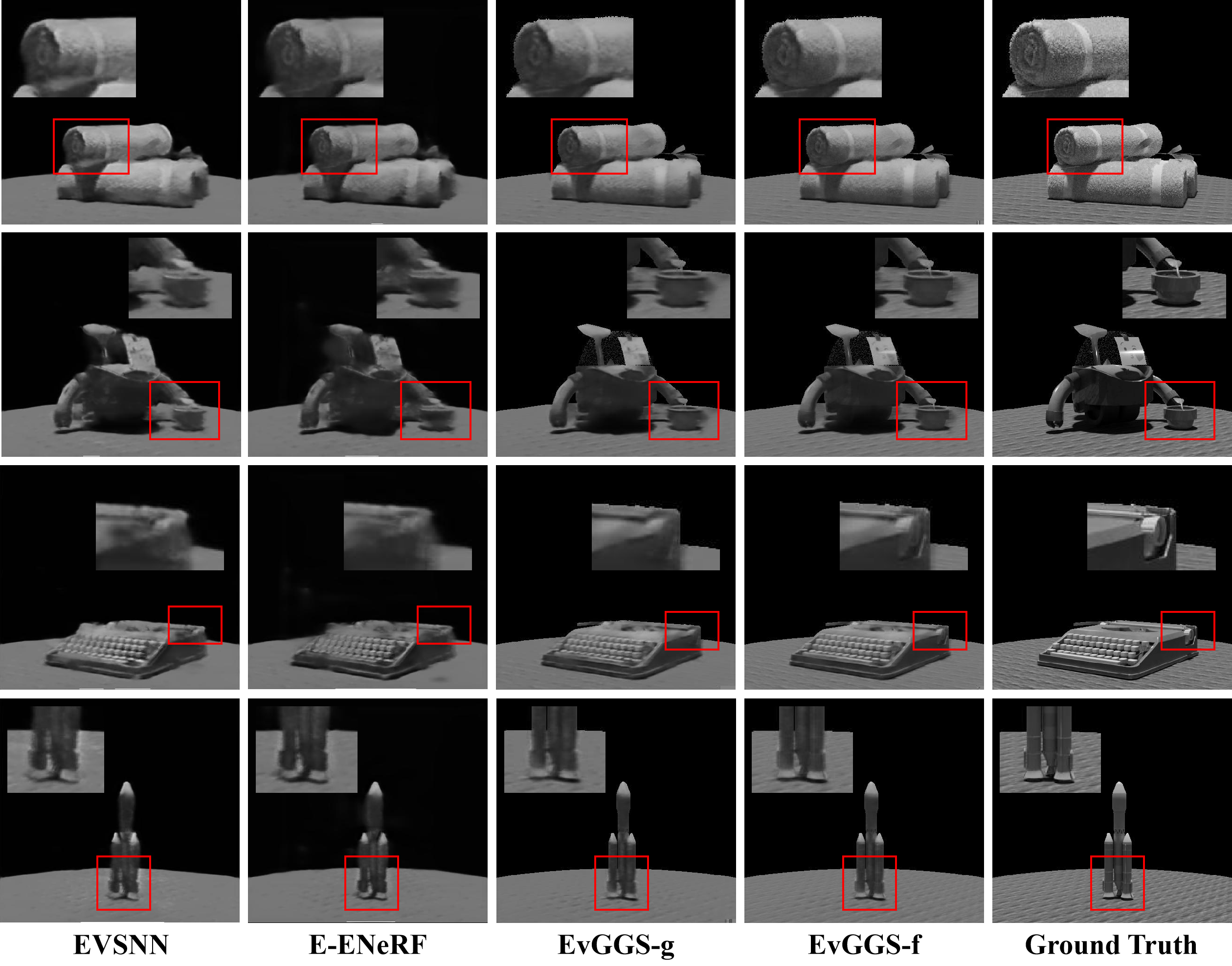}
    \vspace{-4mm}
    \caption{Qualitative comparisons on realistic event dataset.} 
    \label{fig: ev3dr}
\end{figure*}
The quantitative and qualitative experiment results of Ev3D-R can be found in Table.\ref{table: ev3dr} and Fig.\ref{fig: ev3dr} respectively. Here we show some randomly selected visual results. It can be seen that the proposed methods reconstruct a finer texture in all examples.
\subsection{Ablation Studies}
This subsection demonstrates the impact of different training strategies on model performance in the tasks of novel view synthesis and depth estimation. w/o Joint denotes that we only train the Gaussian regressor with the individually trained and frozen depth estimator and intensity reconstructor.
w/o Cascade means that the input does not contain the feature map of the previous network, but only the event voxel and prediction results from the last modules.  
In Table. \ref{Ablation}, the performance of w/o Joint and w/o Cascade is significantly degraded because submodules hardly benefit from the others in the two settings. Besides, w/o ${L}_I$ and w/o ${L}_D$ represent the corresponding loss variants by removing $L_{I_\theta}$ and $L_{D_\phi}$ respectively. 
Table. \ref{Ablation} demonstrates that the absence of depth supervision during joint training leads to a decline in depth estimation. The w/o $L_I$ results in a lack of constraints for intensity reconstruction, which degrades the performance of the subsequent cascaded Gaussian regressor and adversely affects the other two submodules with varying degrees.

\vspace{-0.25cm}
\section{Conclusion}
We first propose the EvGGS, an event-based 3D reconstruction framework that reconstructs 3D Gaussians from raw event streams and generalizes to unobserved scenes without per-scene training. The framework includes three submodules, namely depth estimator, intensity reconstructor, and 3D Gaussian regressor, they are connected hierarchically in feature space. We propose that collaborative training under the 3DGS framework can inject 3D awareness into the submodules to make them mutually promote. We build a novel event-based 3D dataset with well-calibrated intensity, depth, and mask groundtruth. We experimentally prove that the 3D-aware jointly training pipeline further improves the performance of the three modules, and yields better results than the individually trained model and other baselines. Moreover, the generalizable event-based 3DGS reconstruction framework delivers better results than all counterparts. 
\vspace{-3mm}
\section*{Acknowledgements}
This work is supported by the Guangzhou-HKUST(GZ) Joint Funding Program under Grant No. 2023A03J0682.
\vspace{-3mm}
\section*{Impact Statement}
This work aims to advance the pure event-based 3D reconstruction to release the potential of the brain-inspired camera including high dynamic range, low temporal latency, etc.  This work is the first to use only event data to reconstruct 3D Gaussians in a generalizable way.
{
    \small
    \bibliographystyle{icml2024}
    \bibliography{main}
}

\newpage
\appendix
\onecolumn

\clearpage
\appendix
\section{Dataset and Code}
The dataset guidance including download and retrieval is introduced in \textcolor{blue}{https://github.com/Mercerai/EvGGS/}. We also provide the code demonstration for the users to facilitate direct modifications by users.
\section{Detailed Network Architectures}
\begin{figure}[h]
    \centering
    \includegraphics[width=0.55\linewidth]{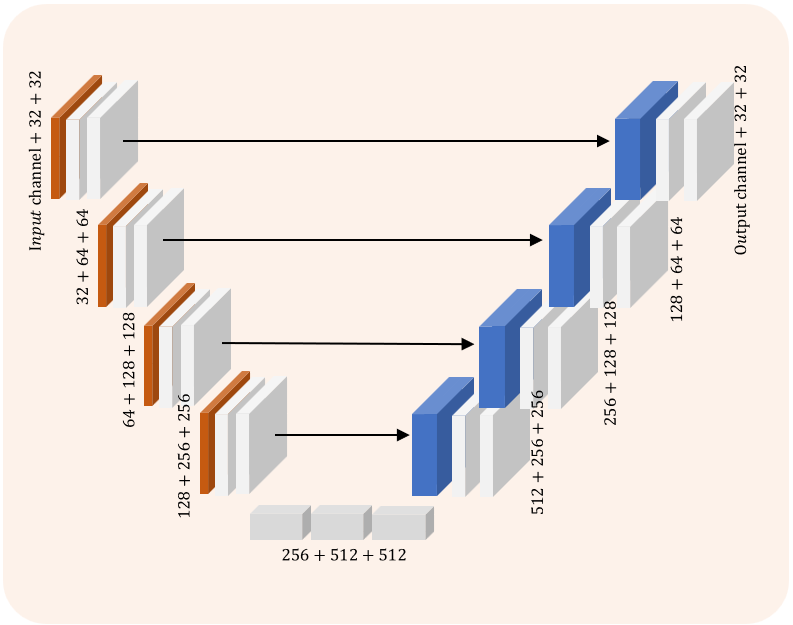}
    \caption{Architecture visualization of the UNet feature extraction network used in depth estimation and intensity reconstruction modules} 
    \label{fig: Unet}
\end{figure}

In this section, we introduce the detailed architectures and parameter selections. The framework includes three cascaded modules, the depth estimator, the intensity reconstructor, and the Gaussian regressor. Among them, the depth estimator and the intensity reconstructor share the same network structure that is a UNet except for their output head. We visualize the network structure in Fig.~\ref{fig: Unet}. Even though they have different input tensors, the input will be transformed into the fixed dimension via a convolution layer with $1 \times 1$ kernels. The depth estimator contains two output heads, one for depth and another for mask, while the intensity reconstructor has a single head to predict the greyscale images. All of them are two independent convolution networks with two hidden layers of $1 \times 1$ kernels. 

\begin{figure}[h]
    \centering
    \includegraphics[width=0.55\linewidth]{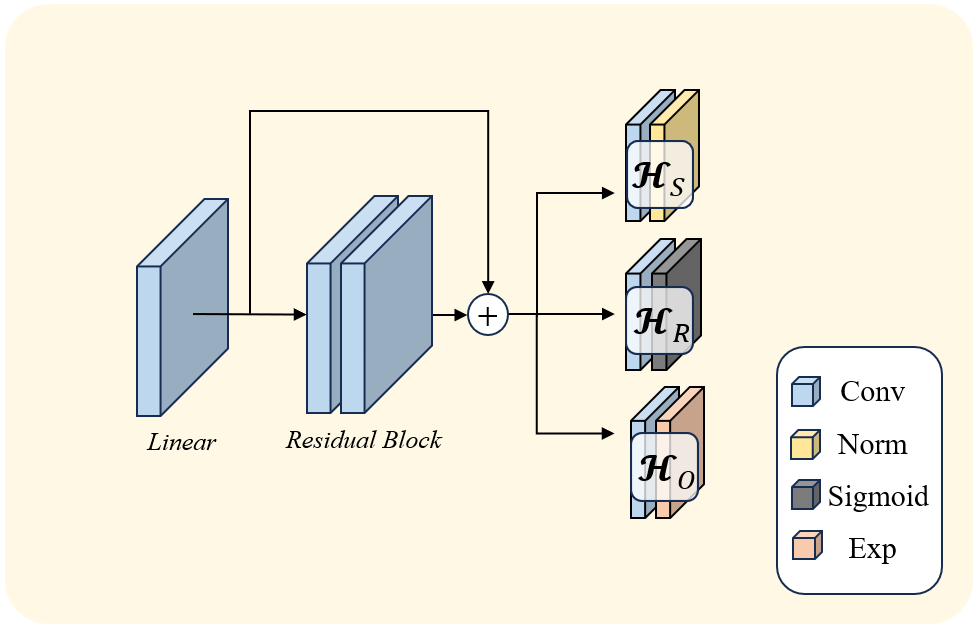}
    \caption{Architecture visualization of the Gaussian regressor network.} 
    \label{fig: EVGSregressor}
\end{figure}

In addition, the Gaussian regressor aims to predict the per-pixel Gaussian parameters, which is a simple convolution network with skip connections, as Fig.~\ref{fig: EVGSregressor} states. We argue that the input of this module includes the depth and intensity map, as well as the high-level features from the last module. The input tensor contains rich high-level semantic meanings thus we do not employ complicated architectures at this step. The "Linear"  in Fig.~\ref{fig: EVGSregressor} refers to a linear projection to transform the input dimension 39 (1 for depth, 1 for intensity, 5 for input event voxel, 32 for the high-level feature from the last module) into the input dimension of the residual block (32). Finally, the output feature is fed to three independent heads to predict the parameter maps with corresponding activations. The three output heads share the same structure as the previously introduced prediction heads in the first two modules. 
\section{Implementation Details}
Our approach and all baselines are trained on a single RTX 3090 GPU by using the Adam optimizer with 1e-5 weighting decay. The initial learning rate is set to 5e-4. We apply the StepLR schedule to adjust the learning rate by multiplying 0.9 every 12000 steps. As the intensity reconstruction module requires the depth feature map as input when training in the collaborative framework, we use a well-trained depth estimator to offer the depth feature map when independently training it. We train 60,000 iterations for the independent training of each submodule in comparisons. Before the collaborative training starts, we only load the checkpoint of the depth estimator at the 60000 step. Then we set the learning rate of the depth estimator to 1e-5, and others remain 5e-4. The entire training process took 9 hours in total. 

\begin{figure*}
    \centering
    \includegraphics[width=0.95\linewidth]{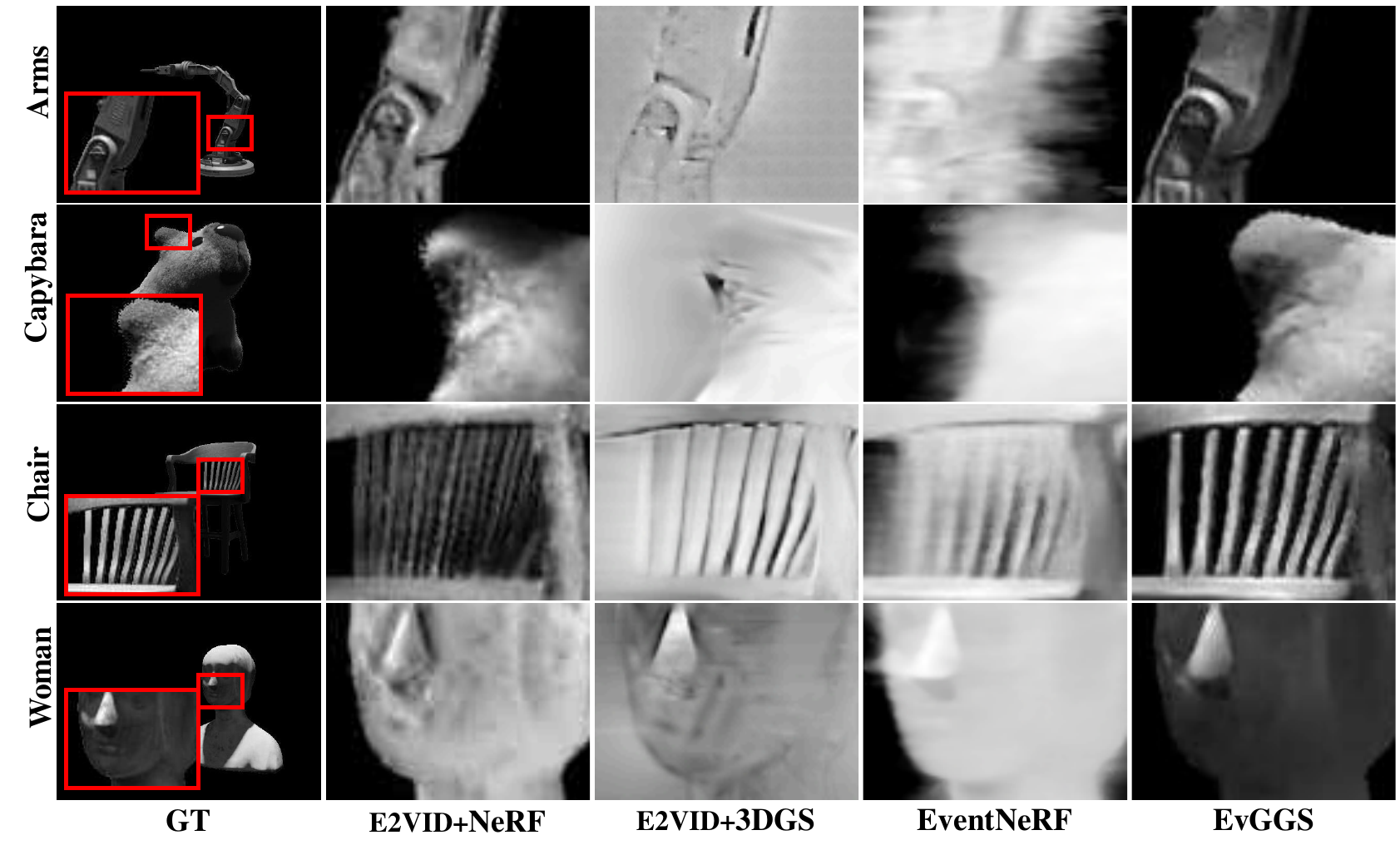}
    \caption{Qualitative comparisons of neural reconstruction with enlarged details.} 
    \vspace{-4mm}
    \label{fig: neural reconstruction with enlarged details}
\end{figure*}
\begin{figure*}
    \centering
    \includegraphics[width=0.95\linewidth]{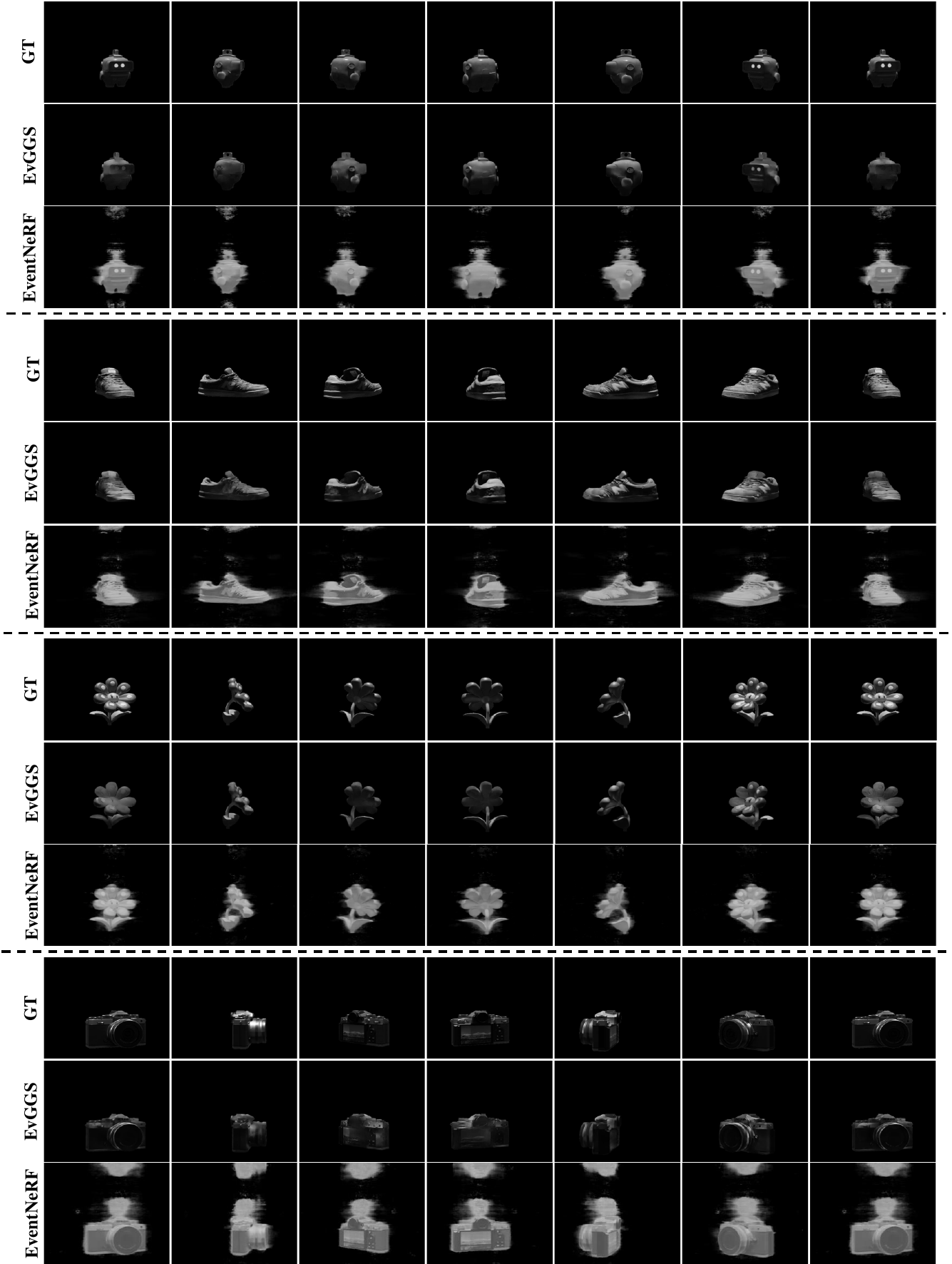}
    \vspace{-8mm}
    \caption{Qualitative comparisons of neural reconstruction with surrounding views.} 
    \label{fig: round views}
\end{figure*}

\section{Visualization of Additional Comparison Results}
We establish additional qualitative comparisons to further showcase our superiorities in the three subtasks including depth estimation, intensity recovery, and novel view synthesis. 

\subsection{Visualization of Reconstructed 3D Objects}
In this section, we want to conduct a more detailed comparison and analysis of the 3D reconstruction results. Figure.\ref{fig: neural reconstruction with enlarged details} gives the large detail boxes of the 3D reconstruction results. Our method consistently achieved the best performance across all test scenes, even when some parts of the scenes had complex geometry and textures or had fewer triggered event points. As can be seen from the first and the third rows, our method is capable of effectively reconstructing complex grid-like and mechanical structures that methods trained on continuous event streams struggle to handle, resulting in sufficiently clear object boundaries. In addition, our method more faithfully restores the original contrast. As the event stream only contains changes in scene luminance, the results of the other three methods all suffer from varying degrees of loss of intensity information. With the help of the jointly trained intensity reconstruction module, our method significantly outperforms other methods in terms of recovering scene contrast.

Figure.\ref{fig: round views} presents surrounding views of 3D reconstruction results for several other scenes. It can be observed that our method does not exhibit any frog or blur from arbitrary viewpoints, thus more closely approximating the groundtruths.

\subsection{Visualization of Recovery Intensity Images}
\begin{figure*}
    \centering
    \includegraphics[width=0.9\linewidth]{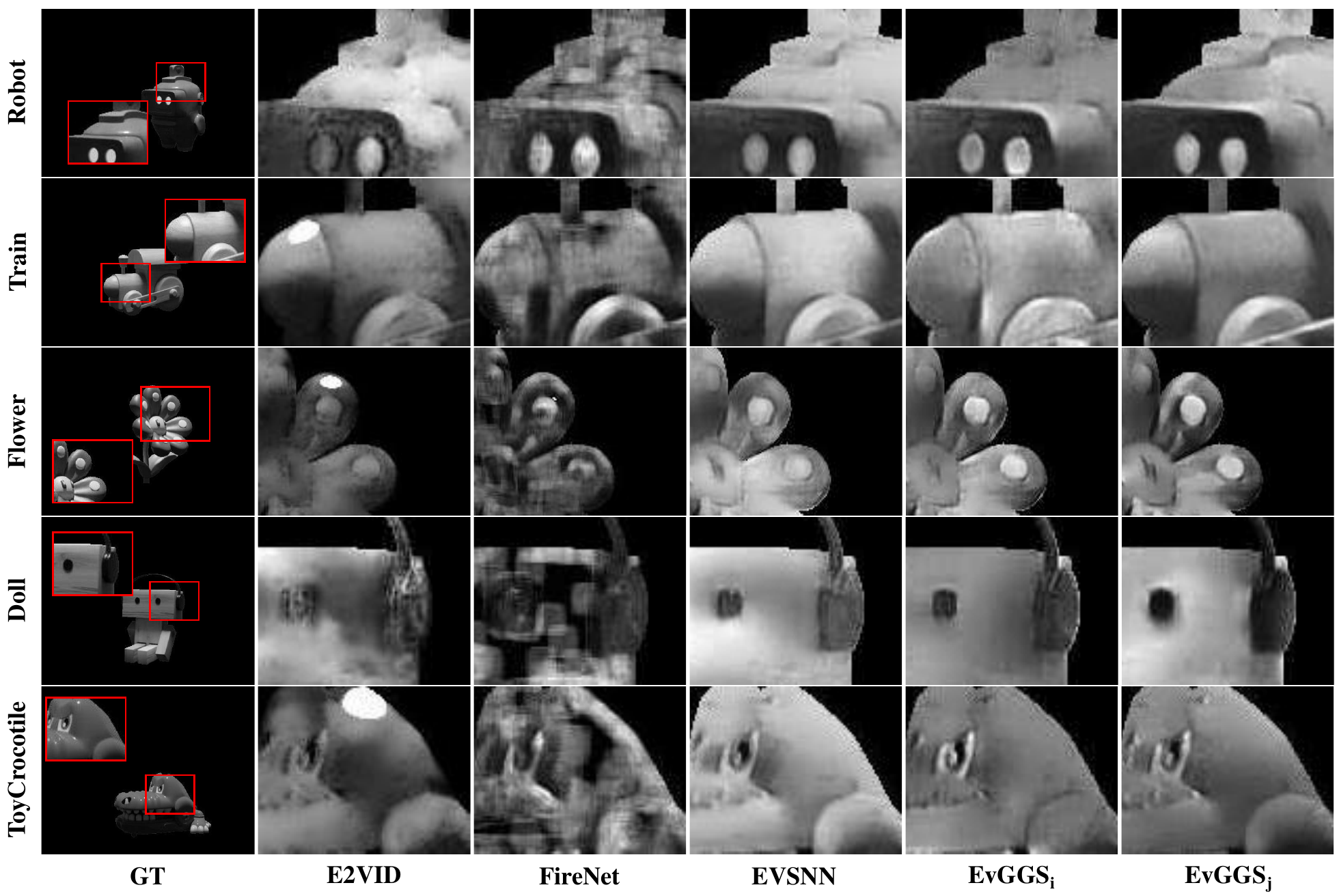}
    \vspace{-4mm}
    \caption{Qualitative comparisons of intensity recovery with enlarged details.} 
    \label{fig: intensity recovery with enlarged details}
    \vspace{-8mm}
\end{figure*}

Figure.\ref{fig: intensity recovery with enlarged details} shows the local details of the intensity recovery results in the enlarged red boxes. FireNet did not recover the intensity values correctly, and it can be observed that there are severe color bleeding effects in all five test scenes. The intensity reconstruction of E2VID is slightly better than FireNet, especially the reconstruction of the ‘Train’ scene, which is quite close to our joint training strategy. However, E2VID incorrectly handled the reflective parts, causing the highlights in the image to turn completely white. Furthermore, the images reconstructed by E2VID also suffer from low contrast and unclear geometric boundaries.

EVSNN performs the best among the methods outside of our joint training strategy, recovering the intensity values well in all scenes except for 'Flower' and 'Doll'. However, it can be observed that EVSNN also has the same issue as E2VID with low contrast. The hue of the scenes reconstructed by EVSNN is noticeably lighter compared to the true intensity map. Moreover, as these three methods rely entirely on the event stream to recover intensity values, all three baseline models are affected in parts where there are fewer triggered event points. Our independent training strategy exhibits a strong sense of flatness in the intensity maps recovered in multiple scenes (Robot, Train, ToyCrocodile), and it fails to distinguish the reflective parts.

\begin{figure*}
    \centering
    \includegraphics[width=0.9\linewidth]{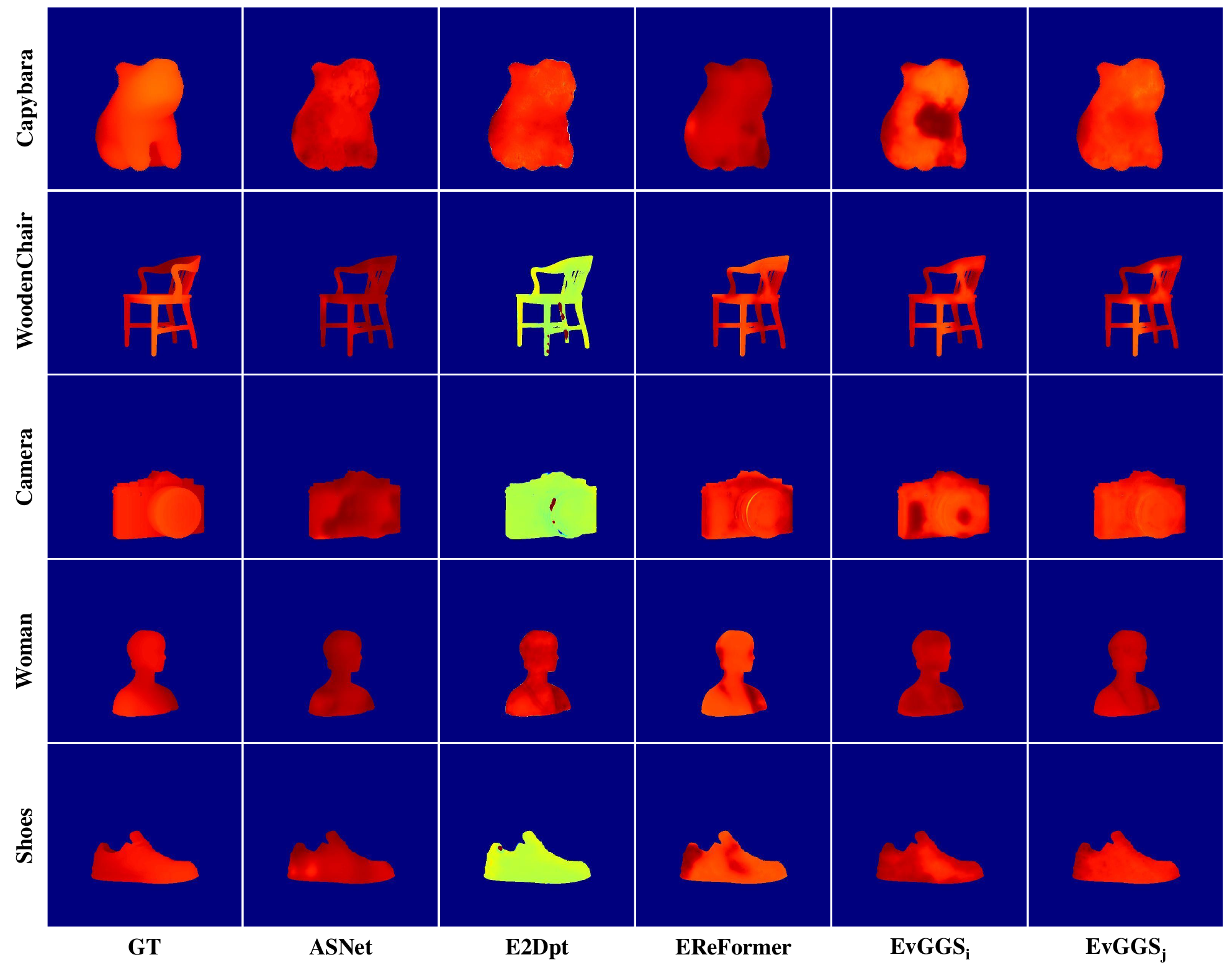}
    \vspace{-5mm}
    \caption{Qualitative comparisons of depth estimation.} 
    \label{fig: other depth estimation scenes}
    \vspace{-8mm}
\end{figure*}
\subsection{Visualizaiton of Depth Map}
In this section, we conduct a detailed analysis of the depth estimation results. Figure.\ref{fig: other depth estimation scenes} shows the qualitative comparison of the depth estimation for other scenes. Our joint training strategy continues to achieve the best depth estimation results in all test scenes. In terms of the Sq.rel metric, E2Dpt is outperformed only by our joint training strategy. However, it exhibits the poorest performance concerning the Abs.rel metric. This indicates that E2Dpt can estimate a continuous and consistent depth map without significant local errors. However, E2Dpt does not obtain the correct depth map, and it gets completely wrong depth ranges in multiple scenarios (Shoes, Camera, Dolls, etc.). Because E2Dpt has a similar network structure to ours, this suggests that using only event spatial-temporal voxels as input cannot extract enough scene information. 

Both ASNet and EReFormer have achieved relatively higher Abs.rel and lower Sq.rel than E2Dpt. As shown in Figure.\ref{fig: Depth} and \ref{fig: other depth estimation scenes}, it can be seen that there are many areas in the depth maps predicted by EReFormer and ASNet where the depth values are discontinuous with the surroundings. This indicates that the utilization of the event frames as input inherently limits the models' capacity for effective 3D scene perception extraction. In addition, due to the relatively complex network structure of ASNet and EReFormer, we adopted the optimized UNet structure as the backbone of our depth estimator to reduce computational load. In other words, our depth estimation module not only has the best depth prediction performance on the test sets, but it is also more suitable for the generalizable 3DGS training pipeline than other methods.

\section{Analysis of Rendering Speed}
Our method benefits from the properties of 3D Gaussians, enabling real-time rendering. As stated in Table~\ref{neural_table}, EventNeRF fails to render in real-time and only produces videos with 0.045 FPS, while the other three models E2VID+ENeRF, E2VID+E3DGS and our EvGGS can interactively produce real-time videos, their FPS are 5, 35, 195 respectively. The rendering speed of E-ENeRF is constrained by the volumetric rendering pipeline that is slower than the Gaussian rasterization. E3DGS delivers the highest FPS because it has been optimized for one scene in advance. However, it still requires retraining when one adopts it to new scenes. Our method needs to recalculate the Gaussian point cloud from scratch from the input data for each rendering time. This is the primary reason resulting in the difference in the rendering speed when compared to the original 3DGS. Nevertheless, by precomputing the Gaussian point cloud and retaining it in memory, we eliminate the need for repetitive computing. Consequently, the subsequent process involves merely rasterization. Under such a precomputation paradigm, our approach is capable of reaching an equivalent rendering speed of 195 FPS. The qualitative comparison results of training and rendering results are shown in Table.\ref{table: speed}. The EvGGS includes three hierarchical stages including intensity reconstruction, depth estimation, and 3DGS regression and rendering. The rendering time can be considered the sum of all previous modules’ inference times. Moreover, we also test the other two event-based 3D reconstruction baselines, i.e. E-ENeRF and EventNeRF. It is noted that the EventNeRF can only optimize on a single scene, thus we only report the time of per-scene optimization. Even though our method includes three modules, the overall training speed and inference speed are still significantly faster than the other two models. Our model can meet the requirement of real-time rendering.
\begin{table*}
\caption{Comparisons of training and rendering speeds.}
\renewcommand{\arraystretch}{1}
\centering
\setlength\tabcolsep{1mm}{
\begin{tabular}{lcccccc}
\hline\hline
& Stage 1 & Stage 2  &Stage 3 & EvGGS-Total   & E-ENeRF & EventNeRF   \\ \hline
Training time    & 4.5h   & 2h   & 5.5h & 12h &15.5h  & 24h  \\
Rendering time    & 0.058s     & 0.056s & 0.118s  & 0.232s & 0.827s  & 7.6s\\
\hline\hline
\end{tabular}}
\label{table: speed}
\vspace{-6mm}
\end{table*}
\section{Additional Experiments}
\subsection{The Data Collection Pipeline to Obtain The Realistic Dataset: EvGGS-R}
In this subsection, We briefly introduce our data collection pipeline. This realistic dataset is captured by the DVXplore event camera. First, we render RGB images and generate corresponding depth groundtruth via Blender, this step is similar to the synthetic dataset. Then we display the videos for these objects on a high fresh rate screen in our work studio with constant low lighting conditions. Meanwhile, we use the DVXplore to continuously capture the screen to obtain the corresponding realistic event stream. Before collecting data, we have already well-calibrated these devices to ensure high-level data association by using chessboard calibration and image, depth, and event frame alignment. 
By using the data collection pipeline, we can obtain a realistic event dataset with corresponding images, depths, and 3D model labels as well. Moreover, The distribution of events captured by the real DVXplore will be more complex, realistic, and disordered, which places higher demands on models. 

\subsection{Qualitative Experiments on Large Scale Scenes}
To show the potential and generalization of the proposed approach, we evaluate our methods on the large scene dataset, MipNeRF 360 \cite{barron2022mipnerf}. We test our method on the Bicycle and Garden scenes. We convert the original large-scale RGB images into event frames by the V2E simulator\cite{hu2021v2e} and evaluate our model and other event-based baselines on that. The quantitative evaluation results are shown in Table.\ref{Tab : unbound}.

\begin{figure*}
    \centering
    \includegraphics[width=0.9\linewidth]{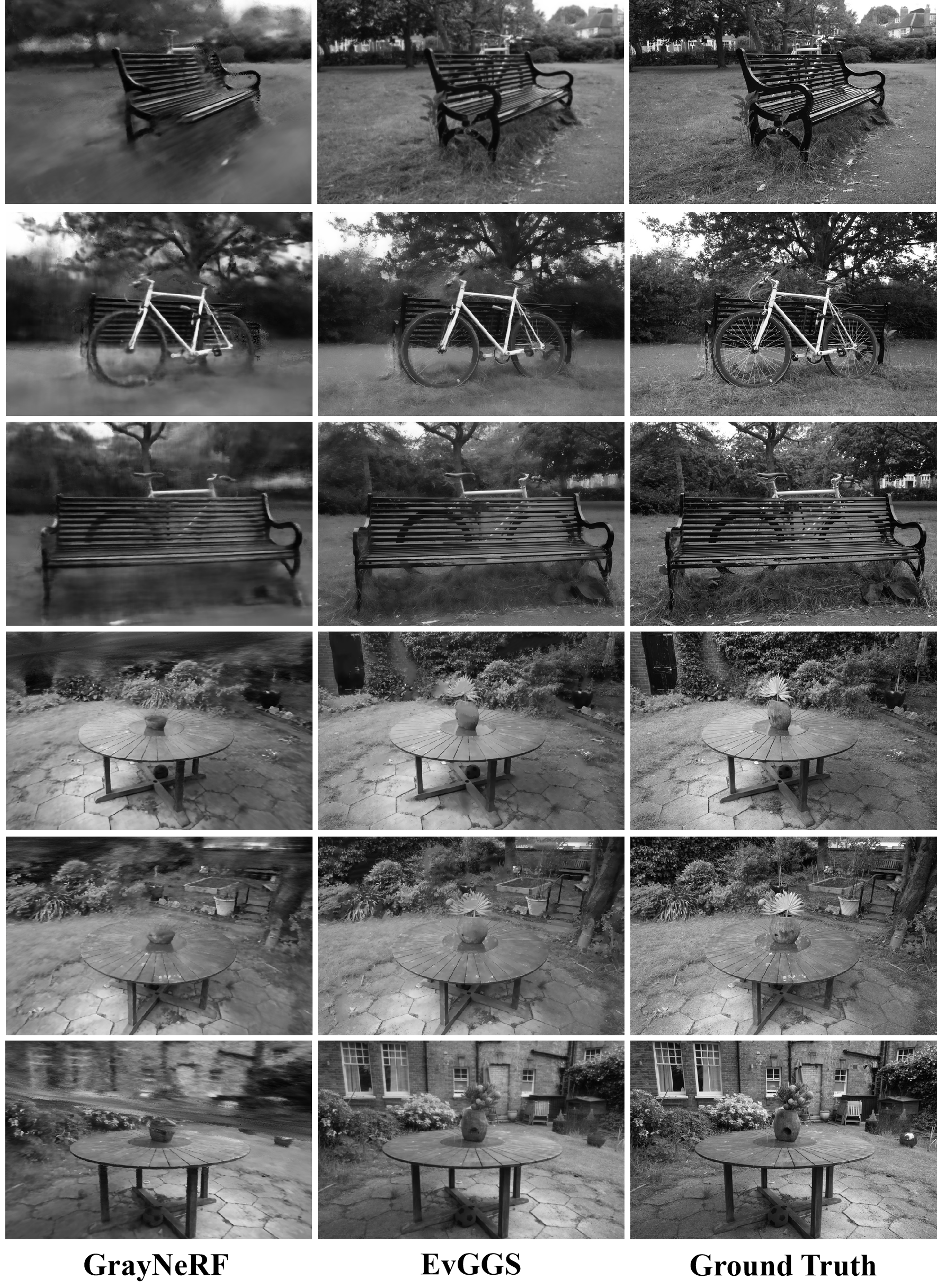}
    \vspace{-5mm}
    \caption{Qualitative comparisons on Mip360.} 
    \label{fig: mip360}
    \vspace{-8mm}
\end{figure*}

In this table, all methods are trained on the synthetic dataset, fine-tuned on 16 views of the MipNeRF 360 dataset, and tested on the other test views of the realistic dataset. It is seen that our method delivers better results than others, which indicates our method can generalize to realistic event data with a small sim2real gap. Moreover, if we finetune our method on 32 distinct views, further improvement will be observed. The visualization results are shown in Fig.\ref{fig: mip360}.

We compare our method with GrayNeRF and E-ENeRF. GrayNeRF means we trained the original NeRF by the grayscale images converted from the RGB images. We optimize the GrayNeRF from scratch because it is a per-scene optimization method. 
\begin{table*}
\caption{Quantitative Results on Unbounded and Large Scenes.}
\renewcommand{\arraystretch}{1.2}
\centering
\setlength\tabcolsep{1.4mm}{
\begin{tabular}{l|ccc|ccc}
\hline\hline
&\multicolumn{3}{c|}{\textbf{Bicycle}} &\multicolumn{3}{c}{\textbf{Garden}}\\ 
&GrayNeRF &E-ENeRF &EvGGS &GrayNeRF &E-ENeRF &EvGGS\\
\hline
PSNR↑    & 19.84 & 18.98   & \textbf{22.75}   & 21.03 & 19.62  &\textbf{24.08}  \\
SSIM↑    & 0.451 & 0.482   & \textbf{0.760}   & 0.677 & 0.490  &\textbf{0.797}  \\
LPIPS↓   & 0.492 & 0.459   & \textbf{0.413}   & 0.435 & 0.511  &\textbf{0.398}  \\
\hline\hline
\end{tabular}}
\label{Tab : unbound}
\vspace{-6mm}
\end{table*}
Even though the GrayNeRF is directly trained on the groundtruth images within a per-scene optimization manner, it fails to produce clear and sharp details. In contrast, the E-ENeRF and EvGGS only need to receive the raw event stream. E-ENeRF only reconstructs super-blur images. However, our EvGGS successfully synthesizes clear backgrounds and sharp edges and yields the best PSNR, SSIM, and LPIPS overall scenes.

\subsection{Additional Experiments on EventNeRF Dataset}
\begin{table*}
\caption{Quantitative results on unbounded and large scenes.}
\renewcommand{\arraystretch}{1.2}
\centering
\setlength\tabcolsep{2.5mm}{
\begin{tabular}{l|ccc|ccc}
\hline\hline
&\multicolumn{3}{c|}{\textbf{EventNeRF}} &\multicolumn{3}{c}{\textbf{EvGGS}}\\ 
&PSNR↑ &SSIM↑ &LPIPS↓ &PSNR↑ &SSIM↑ &LPIPS↓\\
\hline
Drums    & 27.43 & 0.91   & 0.07   & 28.58 & 0.92  & 0.07  \\
Ship    & 25.84 & 0.89   & 0.13   & 29.77 & 0.96  & 0.06  \\
Chair    & 30.62 & 0.94   & 0.05   & 31.43 & 0.93  & 0.05  \\
Ficus    & 31.94 & 0.94   & 0.05   & 32.16 & 0.93  & 0.05  \\
Mic    & 31.78 & 0.96   & 0.03   & 32.61 & 0.97  & 0.02  \\
Hotdog    & 30.26 & 0.94   & 0.04   & 31.29 & 0.95  & 0.04  \\
Material    & 24.10 & 0.94   & 0.07   & 29.15 & 0.96  & 0.04  \\
Lego    & 28.85 & 0.93   & 0.06   & 30.71 & 0.95  & 0.05  \\
\hline\hline
\end{tabular}}
\label{Tab : eventnerf}
\vspace{-2mm}
\end{table*}
we additionally evaluate the proposed approach on the dataset proposed by EventNeRF. EventNeRF includes two datasets, one is the colored NeRF dataset, another is the real dataset. Since the real dataset does not include the image groundtruth to compute metrics, we only test our method on the colored NeRF dataset. This dataset uses the synthetic colored event streams, thus we add a tune mapping function at the end of our model, which is a similar way to that of the EventNeRF. The results are shown in Table.\ref{Tab : eventnerf}. It can be observed that our method delivers better performance than EventNeRF on the dataset. The EvGGS achieves higher PSNR, LPIPS, and SSIM by a distinct margin than EventNeRF.

\section{Limitations}
Although the proposed collaborative learning framework largely improves the performance of the three subtasks, some aspects can still be addressed in the future. First, to accommodate event data, we replaced the spherical harmonics in the original 3DGS with a single intensity value, which reduces its capability to model view-dependent effects. This might be solved
 by displaying and modeling the lighting direction through a separate network pathway. Second, this method cannot effectively reconstruct specular metals. 

\end{document}